\documentclass[10pt,twocolumn,letterpaper]{article}

\usepackage{iccv}
\usepackage{times}
\usepackage{epsfig}
\usepackage{graphicx}
\usepackage{amsmath}
\usepackage{amssymb}
\usepackage{color}
\usepackage{xcolor}

\usepackage[ruled]{algorithm}
\usepackage{algpseudocode}
\usepackage{multirow}
\usepackage{booktabs}
\usepackage{subcaption}
\usepackage[subtle]{savetrees}

\newcommand{\doublecheck}[1]{\textcolor{black}{#1}}    
\newcommand{\keypoint}[1]{\vspace{0.1cm}\noindent\textbf{#1}\quad}   
\newcommand{\cut}[1]{}
\usepackage[pagebackref=true,breaklinks=true,letterpaper=true,colorlinks,bookmarks=false]{hyperref}

\iccvfinalcopy 

\def\iccvPaperID{1488} 

\ificcvfinal\fi
\begin{document}

\title{Episodic Training for Domain Generalization}

\author{Da Li$^{1,2}$, Jianshu Zhang$^{3}$, Yongxin Yang$^{2}$ \\
Cong Liu$^4$, Yi-Zhe Song$^2$ and Timothy M. Hospedales$^{1,2,5}$\\
\\
$^1$Samsung AI Center, Cambridge \quad
$^2$SketchX, CVSSP, University of Surrey \\
$^3$University of Science and Technology of China \quad
$^4$iFlytek Research \quad
$^5$University of Edinburgh \\
{\tt\small da.li1@samsung.com, \{yongxin.yang, y.song\}@surrey.ac.uk, xysszjs@mail.ustc.edu.cn} \\
{\tt\small congliu2@iflytek.com, t.hospedales@ed.ac.uk} 
}

\maketitle

\begin{abstract}
Domain generalization (DG) is the challenging and topical problem of learning models that generalize to novel testing domains with different statistics than a set of known training domains. The simple approach of aggregating data from all source domains and training a single deep neural network end-to-end on all the data provides a surprisingly strong baseline that surpasses many prior published methods. In this paper we build on this strong baseline by designing an episodic training procedure that trains a single deep network in a way that exposes it to the domain shift that characterises a novel domain at runtime. Specifically, we decompose a deep network into feature extractor and classifier components, and then train each component by simulating it interacting with a partner who is badly tuned for the current domain. This makes both components more robust, ultimately leading to our networks producing state-of-the-art performance on three DG benchmarks. Furthermore, we consider the pervasive workflow of using an ImageNet trained CNN as a fixed feature extractor for downstream recognition tasks. Using the Visual Decathlon benchmark, we demonstrate that our episodic-DG training improves the performance of such a general purpose feature extractor by explicitly training a feature for robustness to novel problems. This shows that DG training can benefit standard practice in computer vision.
\end{abstract}

\section{Introduction}

 Machine learning methods often degrade rapidly in performance if they are applied to domains with very different statistics to the data used to train them. This is the problem of domain shift, which domain adaptation (DA) aims to address in the case where some labelled or unlabelled data from the target domain is available for adaptation \cite{shai2006nipsdomainadaptation,tzeng2014deep,long2015learning,ganin2015unsupervised,mslongnips2016,bousmalis2016domain}; and domain generalisation (DG) aims to address in the case where no adaptation to the target problem is possible \cite{muandet2013domaingeneralization,Ghifary2015mtae,Li2018MLDG,shankar2018generalizing} due to lack of data or computation. DG is a particularly challenging problem setting, since explicit training on the target is disallowed; yet it is particularly valuable due to its lack of assumptions. For example, it would be valuable to have a domain-general visual feature extractor that performs well `out of the box' as a representation for any novel problem, even without fine-tuning.
 

The significance of the DG challenge has led to many studies in the literature. These span robust feature space learning \cite{muandet2013domaingeneralization,Ghifary2015mtae}, model architectures that are purpose designed to enable robustness to domain shift \cite{Khosla12undobias,Xu2014lre,Li2017dg} and specially designed learning algorithms for optimising standard architectures \cite{shankar2018generalizing,Li2018MLDG} that aim to fit them to a more robust minima. Among all these efforts, it turns out  that the naive approach \cite{Li2017dg} of aggregating all the training domains' data together and training a single deep network end-to-end is very competitive with state-of-the-art, and better than many published methods -- while simultaneously being much simpler and faster than more elaborate alternatives. In this paper we aim to build on the strength and simplicity of this simple data aggregation strategy, but improve it by designing an episodic training scheme to improve DG.

The paradigm of episodic training  has recently been popularised in the area of few-shot learning \cite{finn2017model,ravi2016optimization,snell2017prototypicalNets}. In this problem, the goal is to use a large amount of background source data, to train a model that is capable of few-shot learning when adapting to a novel target problem. However despite the data availability, training on all the source data would not be reflective of the target few-shot learning condition. So in order to train the model in a way that reflects how it will be tested, multiple few-shot learning training \emph{episodes} are setup among all the source datasets \cite{finn2017model,ravi2016optimization,snell2017prototypicalNets}. 

How can an episodic training approach be designed for domain generalisation? Our insight is that, from the perspective of any layer $l$ in a neural network, being exposed to a novel domain at testing-time is experienced as that layer's neighbours $l-1$ or $l+1$ being badly tuned for the problem at hand. That is,  neighbours provide input to the current layer (or accept output from it) with different statistics to the current layer's expectation. Therefore to design episodes for DG, we should expose layers to neighbours that are untrained for the current domain. If a layer can be trained to perform  well in this situation of badly tuned neighbours, then its robustness to domain-shift has increased. 

To realise our episodic training idea, we break networks up into feature extractor and classifier modules and train them with our episodic framework. This leads to more robust modules that together obtain state-of-the-art results on several DG benchmarks. Our approach benefits from end-to-end learning, while being model agnostic (architecture independent), and simple and fast to train; in contrast to most existing DG techniques that rely on non-standard architectures \cite{Li2017dg}, auxiliary models \cite{shankar2018generalizing}, or non-standard optimizers \cite{Li2018MLDG}.  

Finally, we provide a practical demonstration of the value of explicit DG training, beyond the isolated benchmarks that are common in the literature. Specifically, we consider whether DG can benefit the common practitioner workflow of using an ImageNet \cite{russakovsky2015ilsvrc} pre-trained CNN as a feature extractor for novel tasks and datasets. The standard (homogeneous) DG problem setting assumes shared label-spaces between source and target domain, thus highly restricting its applicability. To benefit the wider computer vision workflow, we go beyond this to \emph{heterogeneous} DG (Table~\ref{tab:vddg-vs-exdg}). That is, to train a feature extractor specifically to improve its robustness in representing novel downstream tasks without fine-tuning. Using the Visual Decathlon benchmark \cite{Rebuffi17}, we show that Episodic training provides an improved representation for novel downstream tasks compared to the standard ImageNet pre-trained CNN.

\section{Related Work}

\keypoint{Multi-Domain Learning (MDL)} MDL aims to learn several domains simultaneously using a single model \cite{Bilen17,Rebuffi17,Rebuffi18,yang2015mdlmtl}. Depending on the problem, how much data is available per domain, and how similar the domains are, multi-domain learning can improve \cite{yang2015mdlmtl} -- or sometimes worsen \cite{Bilen17,Rebuffi17,Rebuffi18} -- performance compared to a single model per domain. MDL is related to DG because the typical setting for DG is to assume a similar setup in that multiple source domains are provided. But that now the goal is to learn how to extract a domain-agnostic or domain-robust model from all those source domains. 
The most rigorous benchmark for MDL is the Visual Decathlon (VD) \cite{Rebuffi17}. We repurpose this benchmark for DG by training a CNN on a subset of the VD domains, and then evaluating its performance as a feature extractor on an unseen disjoint subset of them. We are the first to demonstrate DG at this scale, and in the heterogeneous label setting required for VD.


\keypoint{Domain Generalization} Despite different details, previous DG methods can be divided into a few categories by motivating intuition. Domain Invariant Features: These aim to learn a domain-invariant feature representation, typically by minimising the discrepancy between all source domains -- and assuming that the resulting source-domain invariant feature will work well for the target as well. To this end  \cite{muandet2013domaingeneralization} employed maximum mean discrepancy (MMD), while \cite{Ghifary2015mtae} proposed a multi-domain reconstruction auto-encoder to learn this domain-invariant feature. More recently, \cite{Li2018mmdaae} applied MMD constraints within the representation learning of an autoencoder via adversarial training. Hierarchical Models: These learn a hierarchical set of model parameters, so that the model for each domain is parameterised by a combination of a domain-agnostic and a domain-specific parameter \cite{Khosla12undobias,Li2017dg}. After learning such a hierarchical model structure on the source domains the domain agnostic parameter can be extracted as the model with the least domain-specific bias, that is most likely to work on a target problem. This intuition has been exploited in both shallow \cite{Khosla12undobias} and deep \cite{Li2017dg} settings. Data Augmentation: A few studies proposed data augmentation strategies to synthesise additional training data to improve the robustness of a model to novel domains. These include the Bayesian network \cite{shankar2018generalizing}, which perturbs input data based on the domain classification signal from an auxiliary domain classifier.  Meanwhile, \cite{riccardo_nips18} proposed an adversarial data augmentation method to synthesize `hard' data for the training model to enhance its generalization. Optimisation Algorithms: A final category of approach is to modify a conventional learning algorithm in an attempt to find a more robust minima during training, for example through meta-learning \cite{Li2018MLDG}. Our approach is different to all of these in that it trains a standard deep model, without special data augmentation and with a conventional optimiser. The key idea requires only a simple modification of the training procedure to introduce appropriately constructed episodes. Finally, in contrast to the small datasets considered previously, we demonstrate the impact of DG model training in the large scale VD benchmark.

\keypoint{Neural Network Meta-Learning} Learning-to-learn and meta-learning methods have resurged recently, in particular in few-shot recognition \cite{finn2017model,snell2017prototypicalNets,mishra2017metaTC}, and learning-to-optimize \cite{ravi2016optimization} tasks. Despite signifiant other differences in motivation and methodological formalisations, a common feature of these methods is an episodic training strategy. In few-shot learning, the intuition is that while lot of source tasks and data may be available, these should be used for training in a way that closely simulates the testing condition. Therefore at each learning iteration, a random subset of source tasks and instances are sampled to generate a training episode defined by a random few-shot learning task of similar data volume and cardinality as the model is expected to be tested on at runtime. Thus the model eventually `sees' all the training data in aggregate, but in any given iteration, it is evaluated in a condition similar to a real `testing' condition. In this paper we aim to develop an episodic training strategy to improve domain-robustness, rather than learning-to-learn. While the high-level idea of an episodic strategy is the same, the DG problem and associated episode construction details are completely different.

\section{Methodology}

In this section we will first introduce the basic dataset aggregation method (AGG)  which provides a strong baseline for DG performance, and then subsequently present three episodic training strategies for training it more robustly.

\keypoint{Problem Setting} In the DG setting, we assume that we are given  $n$ source domains $\mathcal{D}=[\mathcal{D}_1, ..., \mathcal{D}_n]$, where $\mathcal{D}_{i}$ is the $i^{th}$\cut{($i \in [1, n]$)} source domain containing data-label pairs {$(\mathbf{x}_i^j, y_i^j)$\footnote{$i$ indicates domain index and $j$ indicates instance number within domain. For simplicity, we will omit $j$ in the following.}}. The goal is to use these to learn a model $f:\mathbf{x}\to y$ that generalises well to a novel testing domain $\mathcal{D}_*$ with different statistics to the training domains, without assuming any knowledge of the testing domain during model learning. 

For \emph{homogeneous} DG, we assume that all the source domains and the target domain share the same label space $\mathcal{Y}_i=\mathcal{Y}_j=\mathcal{Y}_*$, $\forall i,j\in[1,n]$. For the more challenging \emph{heterogeneous} setting, the domains can have different, potentially completely disjoint label spaces $\mathcal{Y}_i\neq\mathcal{Y}_j\neq\mathcal{Y}_*$. We will start by introducing the homogeneous case and discuss the heterogeneous case later.

\keypoint{Architecture} We break neural network classifiers $f:\mathbf{x}\to y$ into a sequence modules. In practice, we use two: A feature extractor $\theta(\cdot)$ and a classifier $\psi(\cdot)$, so that $f(\mathbf{x})=\psi(\theta(\mathbf{x}))$.

\begin{figure}[t]
\centering
\includegraphics[width=1.0\linewidth]{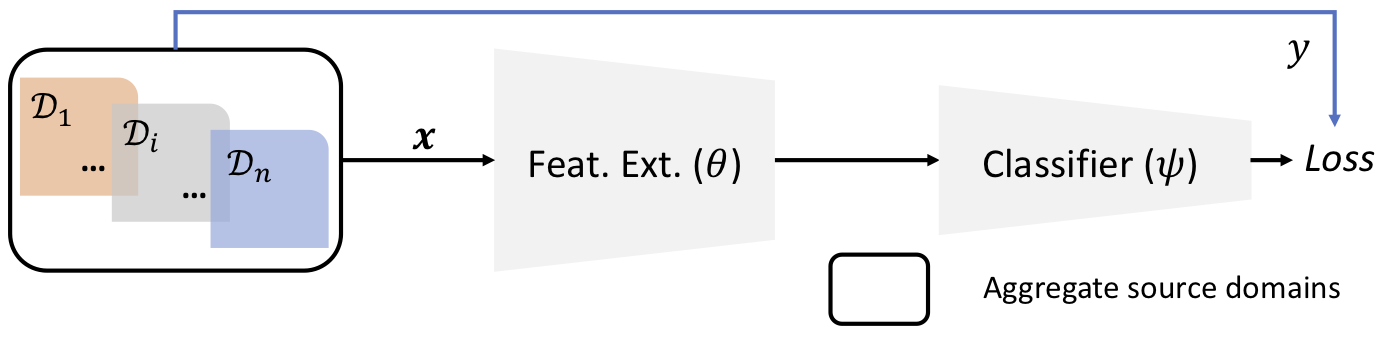}
\vspace{-0.4cm}
\caption{\small Illustration of vanilla domain-aggregation for multi-domain learning. A single model $\psi(\theta(\cdot))$ classifies data from all domains.}
\label{fig-agg}
    \vspace{-0.3cm}
\end{figure}

\subsection{Overview}\label{sec:overview}
\keypoint{Vanilla Aggregation Method} 
A simple approach to the DG problem is to simply aggregate all the source domains' data together, and train a single CNN end-to-end ignoring the domain label information entirely \cite{Li2017dg}. This approach is simple, fast and competitive with more elaborate state-of-the-art alternatives. In terms of neural network modules, it means that both the classifier $\psi$ and the feature extractor $\theta$ are shared across all domains\footnote{At least in the homogeneous case}, as illustrated in Fig.~\ref{fig-agg}, leading to the optimisation:
\begin{equation}
\label{eq:agg}
\begin{aligned}
\underset{\theta, \psi}{\operatorname{argmin}}~ \mathbb{E}_{\mathcal{D}_i\sim\mathcal{D}} \big[ \mathbb{E}_{(\mathbf{x}_i,y_i)\sim \mathcal{D}_i} \big[  \ell(y_i, \psi (\theta(\mathbf{x}_i)) \big] \big]
\end{aligned}
\end{equation}
where $\ell(\cdot)$ is the cross-entropy loss here.


\keypoint{Domain Specific Models}
\begin{figure}[t]
\centering
\includegraphics[width=1.0\linewidth]{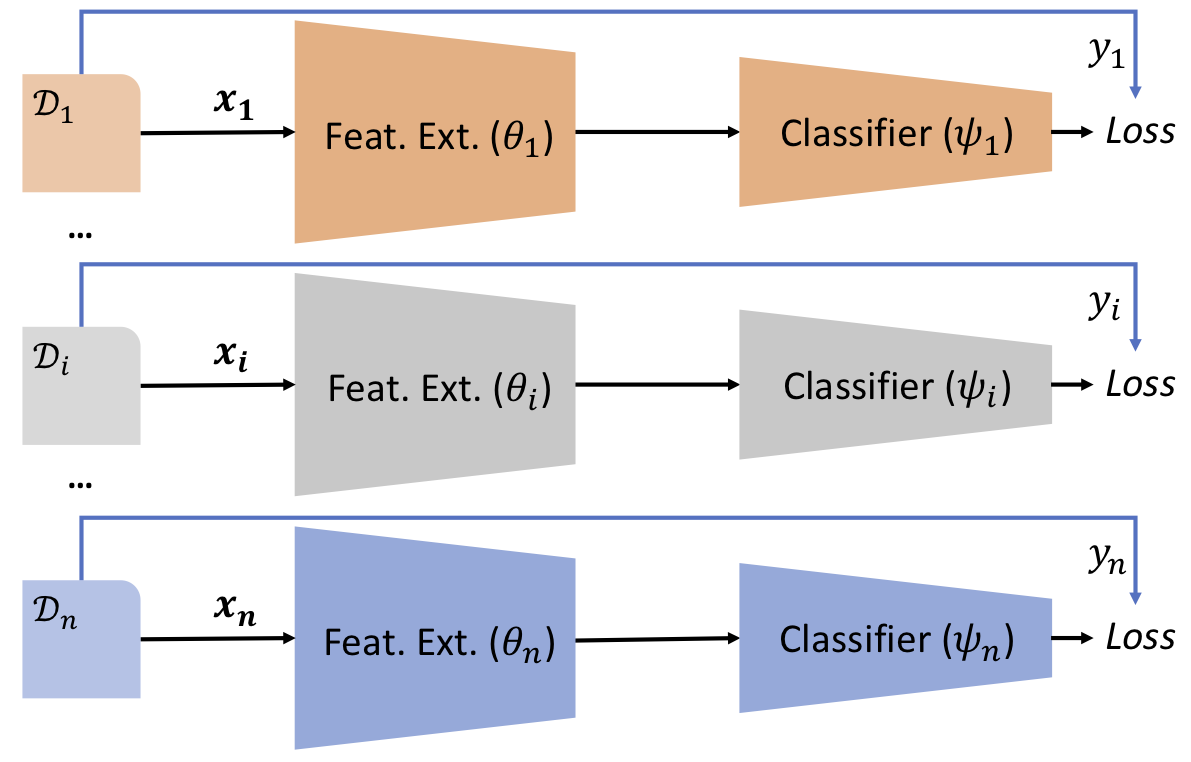}
\vspace{-0.4cm}
\caption{\small Illustration of domain-specific branches. One classifier and feature extractor are trained per-domain.}
\label{fig-dsnn}
    \vspace{-0.3cm}
\end{figure}
Our goal is to improve robustness by exposing individual modules to neighbours that are badly calibrated to a given domain. To obtain these `badly calibrated' components, we also train domain-specific models. As illustrated in Fig.~\ref{fig-dsnn}, this means that each domain $i$ has its own model composed of feature extractor $\theta_i$ and classifier $\psi_i$. Each domain-specific module is only exposed to data of that corresponding domain. To train domain-specific models, we optimise:
\begin{equation}
\label{eq:dsnn}
\begin{aligned}
\underset{[\theta_1,\dots,\theta_n], [\psi_1,\dots,\psi_n]}{\operatorname{argmin}}~ \mathbb{E}_{\mathcal{D}_i\sim\mathcal{D}} \big[ \mathbb{E}_{(\mathbf{x}_i, y_i)\sim \mathcal{D}_i} \big[  \ell(y_i, \psi_i (\theta_i(\mathbf{x}_i)) \big] \big]
\end{aligned}
\end{equation}

\keypoint{Episodic Training}
Our goal is to train a domain agnostic model, as per $\psi$ and $\theta$  in the aggregation method in Eq.~\ref{eq:agg}. And we will design an episodic scheme that makes use of the domain-specific modules as per~Eq.~\ref{eq:dsnn} to help the domain-agnostic model achieve the desired robustness. Specifically, we will generate episodes where each domain agnostic module $\psi$ and $\theta$ is paired with a domain-specific partner that is \emph{mismatched} with the current data being input. So module and data combinations of the form $(\psi,\theta_i,x_{i'})$  and  $(\psi_i,\theta,x_{i'})$ where $i\neq i'$.


\subsection{Episodic Training of Feature Extractor\label{epsisodic-feat}}


To train a robust feature extractor $\theta$, we ask it to learn {features which are robust enough that data from domain $i$ can be processed by a classifier that has never experienced domain $i$ before as shown in Fig.~\ref{fig-agg-cdt}}. To generate episodes according to this criterion, we optimise
\begin{equation}
\begin{aligned}
\label{eq:agg-reg-feat}
\underset{\theta}{\operatorname{argmin}}~
\mathbb{E}_{i,j\sim[1,n], i\neq j} \big[ \mathbb{E}_{(\mathbf{x}_i,y_i)\sim \mathcal{D}_i} \big[  \ell(y_i, \overline{\psi}_j(\theta(\mathbf{x}_i)) \big] \big]
\end{aligned}
\end{equation}
\noindent where $i\neq j$ and  $\overline{\psi}_j$ means that $\psi_j$ is considered constant for the generation of this loss, i.e., it does not receive back-propagated gradients. This gradient-blocking is important, because without it the data $\mathbf{x}_i$  from domain $i$ would `pollute' the classifier $\psi_j$ which we want to retain as being naive to domains outside of $j$.

Thus in this optimisation, only the feature extractor $\theta$ is penalized whenever the classifier $\psi_j$ makes the wrong prediction. That means that, for this loss to be minimised, the shared feature extractor $\theta$ must map data $\mathbf{x}_i$ into a format that a `naive' classifier $\psi_j$ can correctly classify. The feature extractor must learn to help a classifier recognize a data point that is from a domain that is novel to that classifier.


\subsection{Episodic Training of Classifier}
\begin{figure}[t]
\centering
\includegraphics[width=1.0\linewidth]{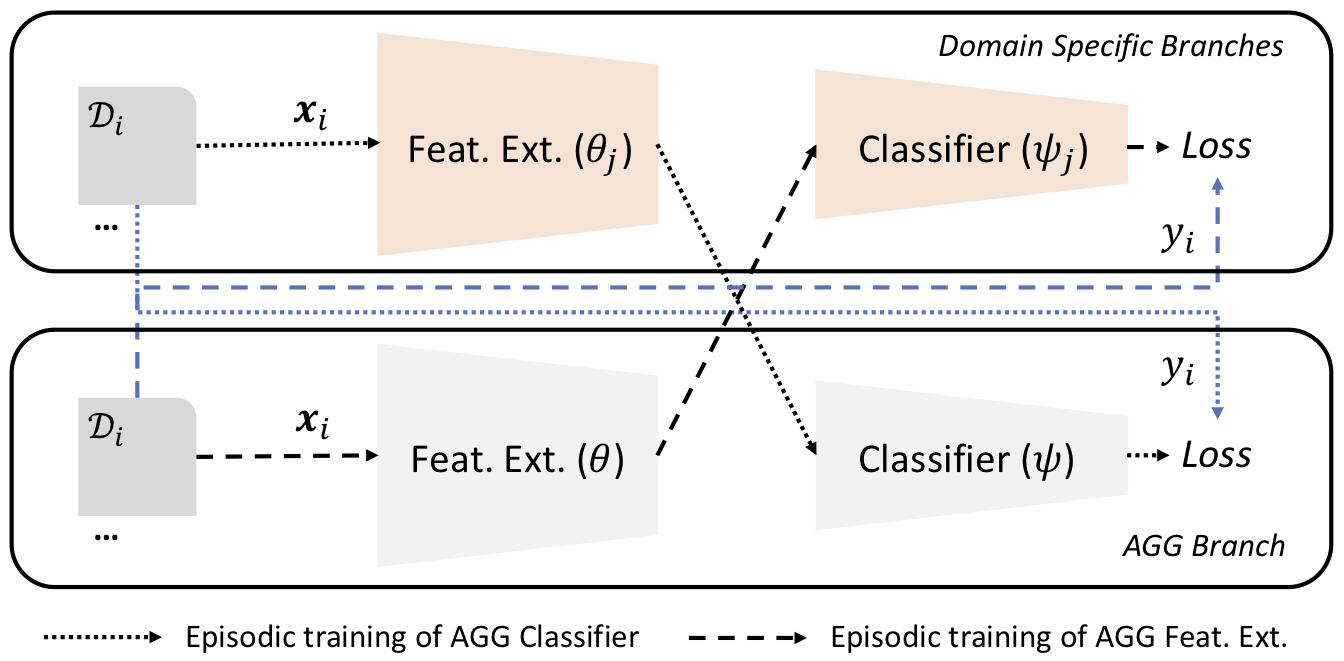}
\vspace{-0.4cm}
\caption{\small Episodic training for feature and classifier regularisation. The shared feature extractor feeds domain specific classifiers. The shared classifier reads domain-specific feature extractors.}
\label{fig-agg-cdt}
    \vspace{-0.3cm}
\end{figure}
Analogous to the above, we can also interpret DG as the requirement that a classifier should be robust enough to classify data even if it is encoded by a feature extractor which has never seen this type of data in the past, as illustrated in Fig.~\ref{fig-agg-cdt}. 
Thus to train the robust classifier $\psi$ we ask it to classify domain $i$ instances $\mathbf{x}_i$ fed through a domain $j$-specific feature extractor $\theta_j$. To generate episodes according to this criterion, we do:
\begin{equation}
\begin{aligned}
\label{eq:agg-reg-clf}
\underset{\psi}{\operatorname{argmin}}~
\mathbb{E}_{i,j\sim[1,n], i\neq j} \big[ \mathbb{E}_{(\mathbf{x}_i,y_i)\sim \mathcal{D}_i} \big[  \ell(y_i, \psi(\overline{\theta}_j(\mathbf{x}_i)) \big] \big]
\end{aligned}
\end{equation}
\noindent where $i\neq j$ and  $\overline{\theta}_j$ means $\theta_j$ is considered constant for generation of the loss here. Similar to the training of the feature extractor module, this operation is important to retain the domain-specificity of feature extractor $\theta_j$. The result is that only the classifier $\psi$ is penalised, and in order to minimise this loss $\psi$ must be robust enough to accept data $\mathbf{x}_i$ that has been encoded by a naive feature extractor $\theta_j$. 


\subsection{Episodic Training by Random Classifier} 

The episodic feature training strategy above is limited to the homogeneous DG setting, since it requires all domains to share  label-space in order to create episodes. But in the heterogeneous scenarios, the shared label-space assumption is not met. We next introduce a novel feature training strategy that is suitable for both homogeneous and heterogeneous label-spaces. 

\begin{figure}[tb]
    \centering
    \includegraphics[width=1.0\linewidth]{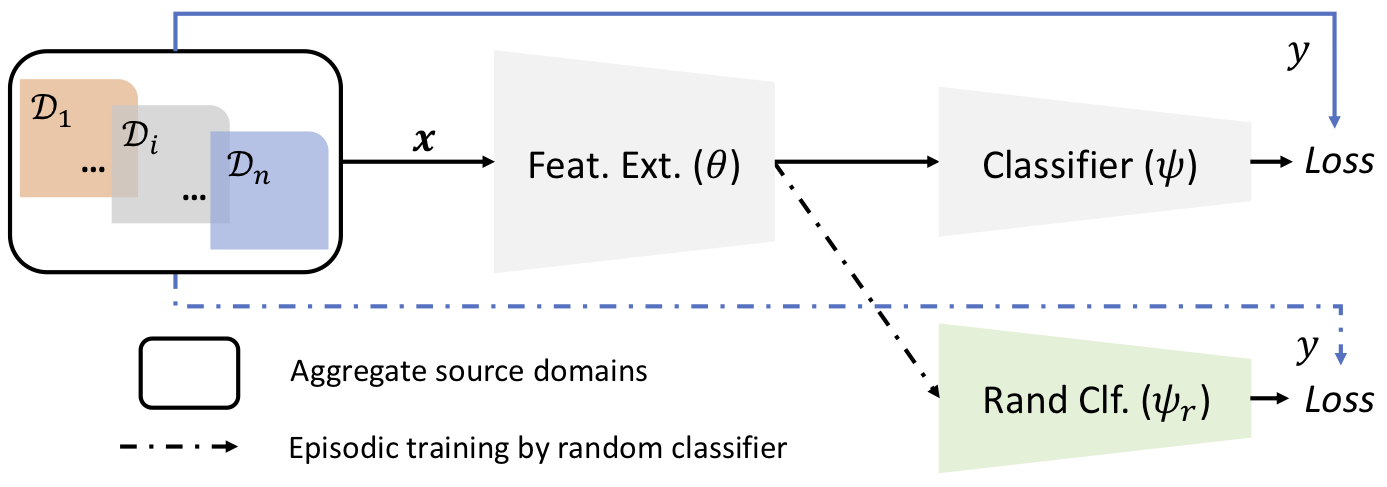}
    \vspace{-0.4cm}
    \caption{\small The architecture of random classifier regularization.}
    \label{fig:agg-fclf}
        \vspace{-0.3cm}
\end{figure}

In Section~\ref{epsisodic-feat}, we introduced the notion of regularising a deep feature extractor by requiring it to support a classifier inexperienced with data from the current domain. Taking this to an extreme, we consider asking the feature extractor to support the predictions of a classifier with \emph{random weights}, as shown in Fig.~\ref{fig:agg-fclf}. To this end, our optimisation task here is:
\begin{equation}
\begin{aligned}
\label{eq:agg-fclf}
\underset{\theta}{\operatorname{argmin}}~
\mathbb{E}_{\mathcal{D}_i\sim\mathcal{D}} \big[ \mathbb{E}_{(\mathbf{x}_i,y_i)\sim \mathcal{D}_i} \big[  \ell(y_i, \overline{\psi}_r(\theta(\mathbf{x}_i)) \big] \big]
\end{aligned}
\end{equation}
\noindent where, $\psi_r$ is a randomly initialised classifier, and $\overline{\psi}_r$ means it is a constant not updated in the optimization. This can be seen as an extreme version of our earlier episodic cross-domain feature extractor training (not only it has not seen any data from domain $\mathbf{x}_i$, but it has not seen any data at all). Moreover, it has the benefit of not requiring a label-space to be shared across all training domains unlike the previous method in Eq.~\ref{eq:agg-reg-feat}. 

Specifically, in Eq.~\ref{eq:agg-reg-feat}, the routing $\mathbf{x}_i \mapsto \theta \mapsto \psi_j$ requires $\psi_j$ to have a label-space matching $(\mathbf{x}_i,y_j)$. But for Eq.~\ref{eq:agg-fclf}, each domain can be equipped with its own random classifier $\psi_r$ with a cardinality matching its normal label-space. This property makes Eq.~\ref{eq:agg-fclf} suitable for heterogeneous domains.


 
\subsection{Algorithm Flow}
Our full algorithm brings together the domain agnostic modules that are our goal to train and the supporting domain-specific modules that help train them (Section~\ref{sec:overview}). We generate episodes according to the three strategies introduced above. Referring the losses in Eq.~\ref{eq:agg}, \ref{eq:dsnn}, \ref{eq:agg-reg-feat}, \ref{eq:agg-reg-clf}, \ref{eq:agg-fclf} as $L_{agg}$, $L_{ds}$, $L_{epif}$, $L_{epic}$, $L_{epir}$, then overall we optimise the objective function:
\begin{equation}
\begin{aligned}
\label{eq:overall}
L_{full}=L_{agg}+L_{ds}+\lambda_1L_{epif}+\lambda_2L_{epic}+\lambda_3L_{epir}
\end{aligned}
\end{equation}

\noindent for parameters $\theta,\phi,\{\theta_i,\psi_i\}_{i=1}^n$. The full pseudocode for the algorithm is given in Algorithm~\ref{alg:episodic}.  It is noteworthy that, in practice,  when training we first warm up the domain-specific branches (sometimes the domain-agnostic one as well) for a few iterations before training both the domain-specific and domain-agnostic modules jointly. {After training, only the domain agnostic modules (of AGG) will be deployed for testing.}

\begin{algorithm}[t]
\caption{Episodic Training for Domain Generalization}\label{alg:episodic}
\begin{algorithmic}[1]
\State \textbf{Input:} $\mathcal{D} = [\mathcal{D}_1, \mathcal{D}_2, \dots, \mathcal{D}_n]$
\State \textbf{Initialise hyper parameters}: $\lambda_1, \lambda_2, \lambda_3, \alpha$
\State \textbf{Initialise model parameters}: 
domain specific modules $\theta_1, ..., \theta_n$ and $\psi_1, ..., \psi_n$; AGG modules $\theta, \psi$; random classifier $\psi_r$
\While{not done training}
\For{$ (\theta_i, \psi_i) \in [(\theta_1, \psi_1),..., (\theta_n, \psi_n)]$}
\State Update $\theta_i:= \theta_i - \alpha \nabla_{\theta_i}(L_{ds})$
\State Update $\psi_i:= \psi_i - \alpha \nabla_{\psi_i}(L_{ds})$
\EndFor
\State Update $\theta := \theta - \alpha \nabla_{\theta}(L_{agg} + \lambda_1 L_{epif} + \lambda_3 L_{epir})$
\State Update $\psi := \psi - \alpha \nabla_{\psi} (L_{agg} + \lambda_2 L_{epic})$
\EndWhile
\State \textbf{Output:} $\theta, \psi$
\end{algorithmic}
\end{algorithm}

\section{Experiments}
\subsection{Datasets and Settings}
\keypoint{Datasets}
We evaluate our algorithm on three different \emph{homogeneous} DG benchmarks and introduce a  novel and larger scale \emph{heterogeneous} DG benchmark. The datasets are: \textbf{IXMAS:} \cite{daniel2006ixmax} is cross-view action recognition task. Two object recognition benchmarks include: \textbf{VLCS} \cite{chen2013vlcs}, which includes images from four famous datasets PASCAL VOC2007 (V) \cite{Everingham10}, LabelMe (L) \cite{russell08labelme}, Caltech (C) \cite{Feifei2004caltech} and SUN09 (S) \cite{Choi2010sun09} and the more recent \textbf{PACS} which has a larger cross-domain gap than VLCS \cite{Li2017dg}. It contains four domains covering Photo (P), Art Painting (A), Cartoon (C) and Sketch (S) images. \textbf{VD}: For the final benchmark we repurpose the Visual Decathlon \cite{Rebuffi17} benchmark to evaluate DG.

\keypoint{Competitors}
We evaluate the following competitors: 
\textbf{AGG} the vanilla aggregation method, introduced in Eq.~\ref{eq:agg}, trains a single model for all source domains.
\textbf{DICA} \cite{muandet2013domaingeneralization} a kernel-based method for learning domain invariant feature representations.
\textbf{LRE-SVM} \cite{Xu2014lre} a SVM-based method, that trains different SVM model for each source domain. For a test domain, it uses the SVM model from the most similar source domain.
\textbf{D-MTAE} \cite{Ghifary2015mtae} a de-noising multi-task auto encoder method, which learns  domain invariant features by cross-domain reconstruction.
\textbf{DSN} \cite{bousmalis2016domain} Domain Separation Networks decompose the sources domains into shared and private spaces and learns them with a reconstruction signal. 
\textbf{TF-CNN} \cite{Li2017dg} learns a domain-agnostic model by factoring out the common component from a set of domain-specific models, as well as tensor factorization to compress the model parameters.
\textbf{CCSA} \cite{motiian2017CCSA} uses semantic alignment to regularize the learned feature subspace.
\textbf{DANN} \cite{ganin2016dann} Domain Adversarial Neural Networks train a feature extractor with a domain-adversarial loss among the source domains. The source-domain invariant feature extractor is assumed to generalise better to novel target domains.
{\textbf{MAML} \cite{finn2017model} The model-agnostic meta-learning method for fast adaptation,  repurposed for DG.}
\textbf{MLDG} \cite{Li2018MLDG} A recent meta-learning based optimization method. It mimics the DG setting by splitting source domains into meta-train and meta-test, and modifies the optimisation to improve meta-test performance.
\textbf{Fusion} \cite{Massimiliano2018ICIP} A method that fuses the predictions from source domain classifiers for the target domain.
\textbf{MMD-AAE} \cite{Li2018mmdaae} A recent method that learns domain invariant feature autoencoding with adversarial training and ensuring that domains are aligned by the MMD constraint.
\textbf{CrossGrad} \cite{shankar2018generalizing} A recent method that uses Bayesian networks to perturb the input manifold for DG.
\textbf{MetaReg} \cite{NIPS2018_metareg} A recent DG method that meta-learns the classifier regularizer.
We note that DANN (domain adaptation) is not designed for DG. However, DANN learns domain invariant features, which is natural for DG. And we found it effective for this problem. Therefore we repurpose it as a baseline.

We call our method as \textbf{Episodic}. We use \textbf{Epi-FCR} to denote our full method with (f)eature regularisation, (c)lassifier regularisation and (r)andom classifier regularisation respectively. Ablated variants such as \textbf{Epi-F} denote feature regularisation alone, etc. Episodic is implemented using PyTorch \footnote{https://github.com/HAHA-DL/Episodic-DG}.

\begin{table*}[tb]
\centering
\scalebox{0.661}{
\begin{tabular}{cc|ccccccccccc}
\toprule
Source  & Target  & DICA \cite{muandet2013domaingeneralization} & LRE-SVM \cite{Xu2014lre} &D-MTAE \cite{Ghifary2015mtae} & CCSA \cite{motiian2017CCSA}&MMD-AAE \cite{Li2018mmdaae} & DANN \cite{ganin2016dann} & MLDG \cite{Li2018MLDG} & CrossGrad \cite{shankar2018generalizing} & MetaReg \cite{NIPS2018_metareg} &AGG & Epi-FCR \\ \midrule
0,1,2,3&4& 61.5 & 75.8& 78.0& 75.8& \textbf{79.1}& 75.0 & 70.7 & 71.6 &74.2& 73.1& 76.9                           \\
0,1,2,4&3& 72.5& 86.9& 92.3& 92.3& 94.5 & 94.1 & 93.6 & 93.8 &94.0& 94.2& \textbf{94.8}                           \\
0,1,3,4&2& 74.7& 84.5& 91.2& 94.5& 95.6 & 97.3 & 97.5 & 95.7 &96.9& 95.7& \textbf{99.0}                           \\
0,2,3,4&1& 67.0& 83.4& 90.1& 91.2& 93.4 & 95.4 & 95.4 &94.2 &97.0& 95.7& \textbf{98.0}                          \\
1,2,3,4&0& 71.4& 92.3& 93.4& \textbf{96.7}& \textbf{96.7} & 95.7 & 93.6 & 94.0 &94.7& 94.4& 96.3                          \\
\midrule
\multicolumn{2}{c|}{Ave.} & 69.4& 84.6& 87.0& 90.1& 91.9& 91.5 & 90.2 &89.9 &91.4& 90.6&  \textbf{93.0}      \\
\bottomrule
\end{tabular}
}
\vspace{-0.3cm}
\caption{\small Cross-view action recognition results (accuracy. \%) on IXMAS dataset. Best result in bold.}
\label{tab:ixmas}
\vspace{-0.1cm}
\end{table*}

\begin{table*}[t]
\centering
\scalebox{0.665}{
\begin{tabular}{cc|ccccccccccc}
\toprule
Source& Target & DICA \cite{muandet2013domaingeneralization}& LRE-SVM \cite{Xu2014lre} & D-MTAE \cite{Ghifary2015mtae}& CCSA \cite{motiian2017CCSA}& MMD-AAE \cite{Li2018mmdaae}& DANN \cite{ganin2016dann}& MLDG \cite{Li2018MLDG} & CrossGrad \cite{shankar2018generalizing} & MetaReg \cite{NIPS2018_metareg} & AGG& Epi-FCR \\
\midrule
L,C,S&V& 63.7& 60.6& 63.9& 67.1& \textbf{67.7}& 66.4 & 67.7 & 65.5 &65.0& 65.4& 67.1\\
V,C,S&L& 58.2& 59.7& 60.1& 62.1& 62.6& 64.0 & 61.3 & 60.0 &60.2& 60.6& \textbf{64.3}\\
V,L,S&C& 79.7& 88.1& 89.1& 92.3& \textbf{94.4}& 92.6 & \textbf{94.4} & 92.0 &92.3& 93.1& 94.1\\
V,L,C&S& 61.0& 54.9& 61.3& 59.1& 64.4& 63.6 & \textbf{65.9} & 64.7 &64.2& 65.8& \textbf{65.9}\\
\midrule
\multicolumn{2}{c|}{Ave.}& 65.7& 65.8& 68.6& 70.2& 72.3& 71.7 & 72.3 & 70.5 &70.4& 71.2& \textbf{72.9}\\
\bottomrule
\end{tabular}
}
\vspace{-0.3cm}
\caption{\small Cross-dataset object recognition results (accuracy. \%) on VLCS benchmark. Best in bold.}
\label{tab:vlcs}
    \vspace{-0.3cm}
\end{table*}

\subsection{Evaluation on \textbf{\textit{IXMAS}} dataset}
\keypoint{Settings} IXMAS contains 11 different human actions. All actions were video recorded by 5 cameras with different views (referred as 0,...,4). The goal is to train an action recognition model on a set of source views (domains), and recognise the action from a novel target view (domain).
We follow \cite{Li2018mmdaae} to keep the first 5 actions and use the same Dense trajectory features as  input. For our method, we follow \cite{Li2018mmdaae} to use a one-hidden layer network with 2000 hidden neurons as our backbone and report the average result of 20 runs. The optimizer is M-SGD with learning rate 1e-4, momentum 0.9, weight decay 5e-5. We use $\lambda_1$=2.0, $\lambda_2$=2.0, and $\lambda_3$=0.5.




\keypoint{Results} From the results in Table \ref{tab:ixmas}, we can see that: (i) The vanilla aggregation method, AGG is a strong competitor compared to several prior published methods, as is DANN, which is newly identified by us as an effective DG algorithm. (ii) Overall our Epi-FCR performs best, improving 2.4\% on AGG, and 1.1\% on prior state-of-the-art MMD-AAE. (iii) Particularly in view 1\&2 our method achieves new state-of-the art performance. 

\subsection{Evaluation on \textbf{\textit{VLCS}} dataset}

\keypoint{Settings} VLCS domains share 5 categories: bird, car, chair, dog and person. We use  pre-extracted DeCAF6 features and follow \cite{motiian2017CCSA} to randomly split each domain into train (70\%) and test (30\%) and do leave-one-out evaluation. We use a 2 fully connected layer architecture with output size of 1024 and 128 with ReLU activation, as per \cite{motiian2017CCSA} and report the average performance of 20 trials. The optimizer is M-SGD with learning rate 1e-3, momentum 0.9 and weight decay 5e-5. We use $\lambda_1$=7.0, $\lambda_2$=5.0, and $\lambda_3$=0.5. 

\keypoint{Results} From the results in Table~\ref{tab:vlcs}, we can see that: (i) The simple AGG baseline is again competitive with many published alternatives, so is DANN.  (ii) Our Epi-FCR method achieves the best performance, improving on AGG by 1.7\% and performing comparably to prior state-of-the-art MMD-AAE and MLDG with  0.6\% improvement over both. 

\subsection{Evaluation on \textbf{\textit{PACS}} dataset}

\keypoint{Settings} PACS is a recent dataset with different object style depictions, and a more challenging domain shift than VLCS, as shown in \cite{Li2017dg}.  This dataset shares 7 object categories across domains, including dog, elephant, giraffe, guitar, house, horse and person. We follow the protocol in \cite{Li2017dg} including the recommended  train and validation split for  fair comparison. We first follow \cite{Li2017dg} in using the ImageNet pretrained AlexNet (in Table~\ref{tab:agg-alex}) and subsequently also use a modern ImageNet pre-trained ResNet-18 (in Table~\ref{tab:agg-resnet-pacs}) as a base CNN architecture. We train our network using the M-SGD optimizer (batch size/per domain=32, lr=1e-3, momentum=0.9, weight decay=5e-5) for 45k iterations when using AlexNet and train our network using the same optimizer (weight decay=1e-4) for ResNet-18\cut{(average results are reported with 10 runs)}. We use $\lambda_1$=2.0, $\lambda_2$=0.05, and $\lambda_3$=0.1 for both settings. We use the official PACS protocol and split \cite{Li2017dg} and rerun MetaReg~\cite{NIPS2018_metareg} on this split, since MetaReg did not release their protocol.


\cut{
\begin{table*}[t]
\centering
\scalebox{0.7}{
\begin{tabular}{cc|cccccccccccc}
\toprule
Source & Target & DICA \cite{muandet2013domaingeneralization} & LRE-SVM \cite{Xu2014lre} &D-MTAE \cite{Ghifary2015mtae}& DSN \cite{bousmalis2016domain} & TF-CNN \cite{Li2017dg} & MLDG \cite{Li2018MLDG} & Fusion \cite{Massimiliano2018ICIP} & DANN \cite{ganin2016dann}  & CrossGrad \cite{shankar2018generalizing} & MetaReg \cite{NIPS2018_metareg} & AGG & Epi-FCR\\
\midrule
C,P,S&A& 64.6&59.7&60.3& 61.1& 62.9& 66.2 & 64.1& 63.2 &  61.0 & 63.5 &63.4 & 64.7\\
A,P,S&C& 64.5&52.8&58.7& 66.5& 67.0& 66.9& 66.8& 67.5 & 67.2 & 69.5 &66.1 & 72.3 \\
A,C,S&P& 91.8 &85.5&91.1& 83.3& 89.5& 88.0& 90.2& 88.1 & 87.6 & 87.4 &88.5 & 86.1\\
A,C,P&S& 51.1&37.9&47.9& 58.6& 57.5& 59.0& 60.1& 57.0 & 55.9 & 59.1 &56.6 & 65.0\\
\midrule
\multicolumn{2}{c|}{Ave.}& 68.0 &59.0&64.5& 67.4& 69.2& 70.0& 70.3& 69.0 &  67.9 &  69.9 &68.7 & 72.0\\
\bottomrule
\end{tabular}}
\vspace{-0.3cm}
\caption{\small Cross-domain object recognition results (accuracy. \%) of different methods on PACS using pretrained AlexNet.}
    \label{tab:agg-alex}
\end{table*}}

\begin{table*}[t]
\centering
\scalebox{0.7}{
\begin{tabular}{cc|ccccccccccc}
\toprule
Source & Target & DICA \cite{muandet2013domaingeneralization} &D-MTAE \cite{Ghifary2015mtae}& DSN \cite{bousmalis2016domain} & TF-CNN \cite{Li2017dg}  & Fusion \cite{Massimiliano2018ICIP} & DANN \cite{ganin2016dann} & MLDG \cite{Li2018MLDG} & CrossGrad \cite{shankar2018generalizing} & MetaReg \cite{NIPS2018_metareg} & AGG & Epi-FCR\\
\midrule
C,P,S&A& 64.6&60.3& 61.1& 62.9& 64.1& 63.2 & \textbf{66.2} &  61.0 & 63.5 &63.4 & 64.7\\
A,P,S&C& 64.5&58.7& 66.5& 67.0&  66.8& 67.5 & 66.9 & 67.2 & 69.5 &66.1 & \textbf{72.3} \\
A,C,S&P& \textbf{91.8} &91.1& 83.3& 89.5&  90.2& 88.1  &88.0 & 87.6 & 87.4 &88.5 & 86.1\\
A,C,P&S& 51.1&47.9& 58.6& 57.5& 60.1& 57.0 & 59.0 & 55.9 & 59.1 &56.6 & \textbf{65.0}\\
\midrule
\multicolumn{2}{c|}{Ave.}& 68.0 &64.5& 67.4& 69.2& 70.3& 69.0& 70.0 &  67.9 &  69.9 &68.7 & \textbf{72.0}\\
\bottomrule
\end{tabular}}
\vspace{-0.3cm}
\caption{\small Cross-domain object recognition results (accuracy. \%) of different methods on PACS using pretrained AlexNet. Best in bold.}
\vspace{-0.4cm}
    \label{tab:agg-alex}
\end{table*}

\keypoint{Results} From the AlexNet results in Table \ref{tab:agg-alex}, we can see that: (i) Our episodic method obtained the best performance on held out domains C and S and comparable performance on A, P domains. (ii) It also achieves the best performance overall, with 3.3\% improvement on vanilla AGG, and at least 1.7\% improvement on prior state-of-the-art methods MLDG \cite{Li2018MLDG}, Fusion \cite{Massimiliano2018ICIP} and MetaReg \cite{NIPS2018_metareg}. 

Meanwhile in Table~\ref{tab:agg-resnet-pacs}, we see that with a modern ResNet-18 architecture, the basic results are improved across the board as expected. However \cut{: (i) While our newly identified DANN manages to improve on the vanilla AGG, further demonstrating their effectiveness on DG. (ii)} our full episodic method maintains the best performance overall, with a 2.4\% improvement on AGG. 

We note here that when using modern architectures like \cite{Szegedy2016inceptionv3, He2016resnet} for DG tasks we need to be careful with batch normalization \cite{Ioffe2015bn}. Batchnorm accumulates statistics of the training data during training, for use at testing. In DG, the source and target domains have domain-shift between them, so different ways of employing batch norm produce different results. We tried two ways of coping with batch norm, one is directly using frozen pre-trained ImageNet statistics. Another  is to unfreeze and accumulate statistics from the source domains. We observed that when training  ResNet-18 on PACS with accumulating the statistics from source domains it produced a worse accuracy than freezing ImageNet statistics ($75.7\%$ vs $79.1\%$).

\begin{table}[t]
\centering
\scalebox{0.53}{
\begin{tabular}{cc|ccccccc}
\toprule
Source & Target & AGG & DANN \cite{ganin2016dann} & MAML~\cite{finn2017model} & MLDG \cite{Li2018MLDG}  & CrossGrad \cite{shankar2018generalizing} & MetaReg \cite{NIPS2018_metareg} & Epi-FCR \\ \midrule
C,P,S& A & 77.6 & 81.3 & 78.3 & 79.5   & 78.7 & 79.5 & \textbf{82.1} \\
A,P,S& C & 73.9 & 73.8 & 76.5 & \textbf{77.3}   & 73.3 & 75.4 & 77.0 \\
A,C,S& P & 94.4 & 94.0 &  \textbf{95.1} & 94.3  & 94.0 & 94.3 & 93.9 \\
A,C,P& S & 70.3 & \textbf{74.3} & 72.6 & 71.5   & 65.1 & 72.2 & 73.0 \\
\midrule
\multicolumn{2}{c|}{Ave.} & 79.1 & 80.8 & 80.6 & 80.7  & 77.8 & 80.4 & \textbf{81.5} \\
\bottomrule
\end{tabular}
}
\vspace{-0.3cm}
\caption{\small Cross-domain object recognition results (accuracy. \%) of different methods on PACS using ResNet-18. Best in bold.}
    \label{tab:agg-resnet-pacs}
      \vspace{-0.3cm}
\end{table}

\begin{figure}[t]
    \centering
    \includegraphics[width=0.49\linewidth]{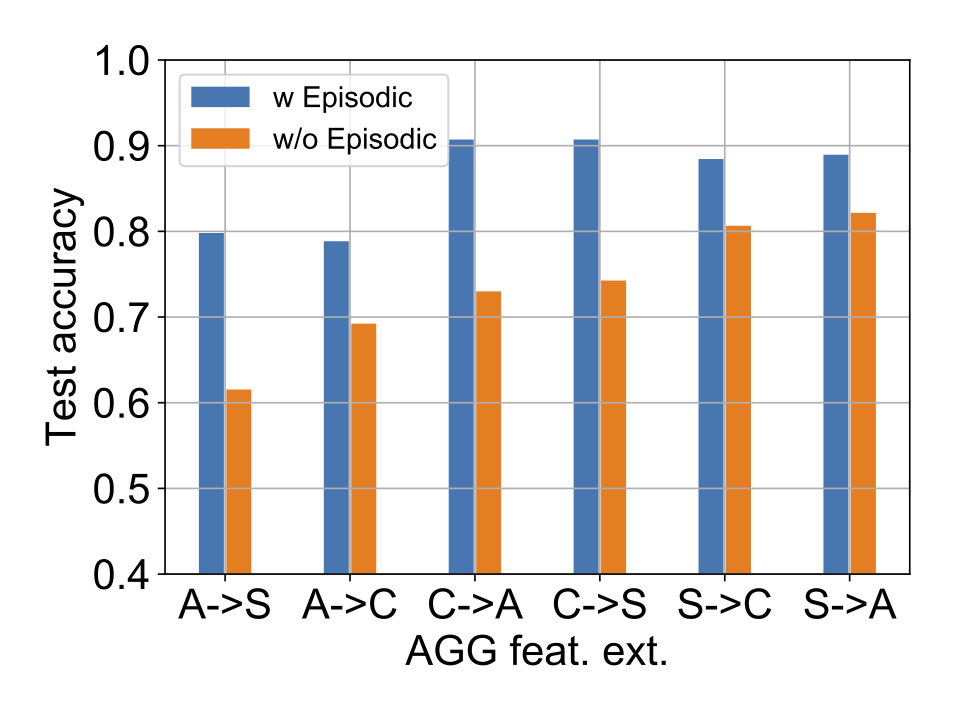}
    \includegraphics[width=0.49\linewidth]{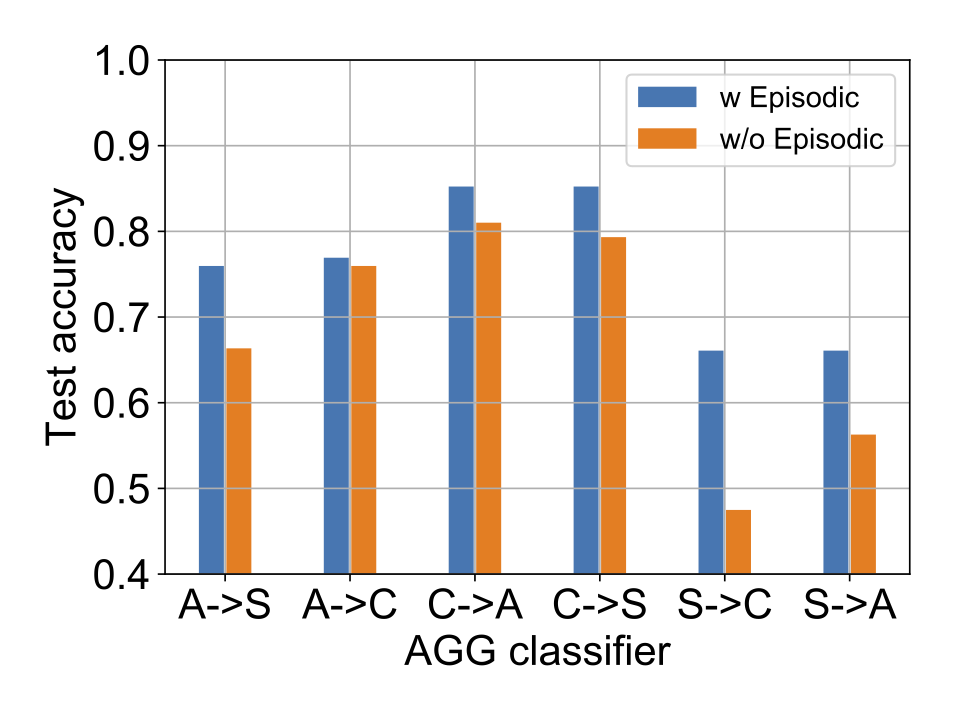}
    \vspace{-0.3cm}
    \caption{\small Cross-domain test accuracy on PACS (AlexNet) with shared feature extractor or classifier. A$\mapsto$C means, feed A data through C-specific module. Eg, left: $\mathbf{x}_A \mapsto \theta \mapsto \psi_{C}$, right: $\mathbf{x}_A \mapsto \theta_C \mapsto \psi$.}
    \label{fig:cross-domain-test-acc}
    \vspace{-0.3cm}
\end{figure}



\subsection{Further Analysis and Insights}
\begin{figure}[t]
    \centering
    \begin{subfigure}[b]{0.4185\linewidth}
    \includegraphics[width=1.0\linewidth]{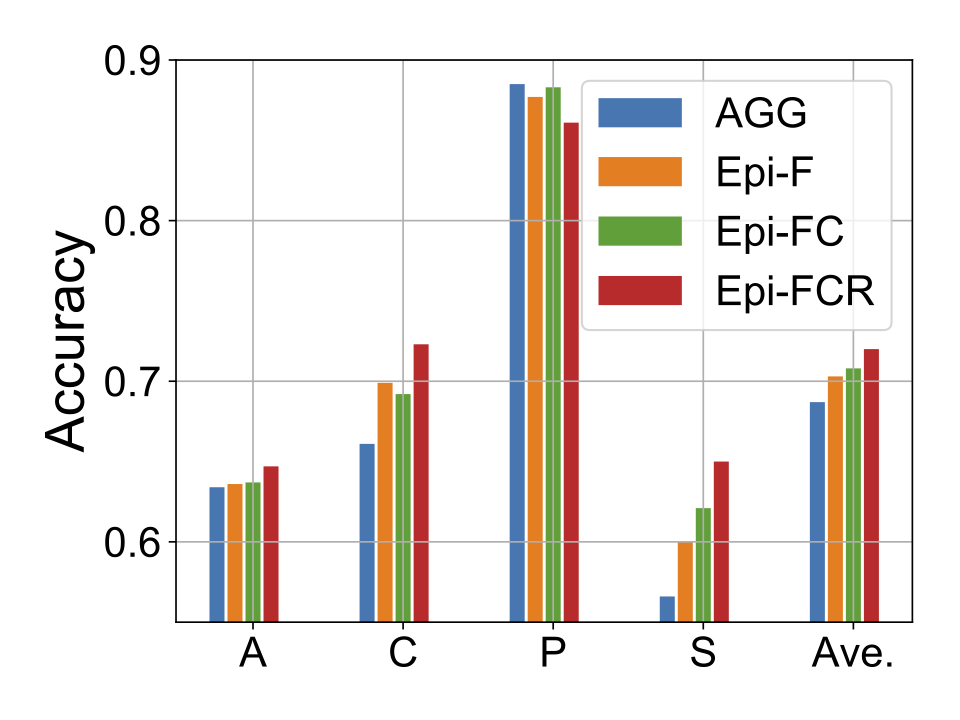}
    \vspace{-0.77cm}
    \caption{}
    \label{fig:ab-pacs}
    \end{subfigure}
    \begin{subfigure}[b]{0.5715\linewidth}
    \includegraphics[width=1.0\linewidth]{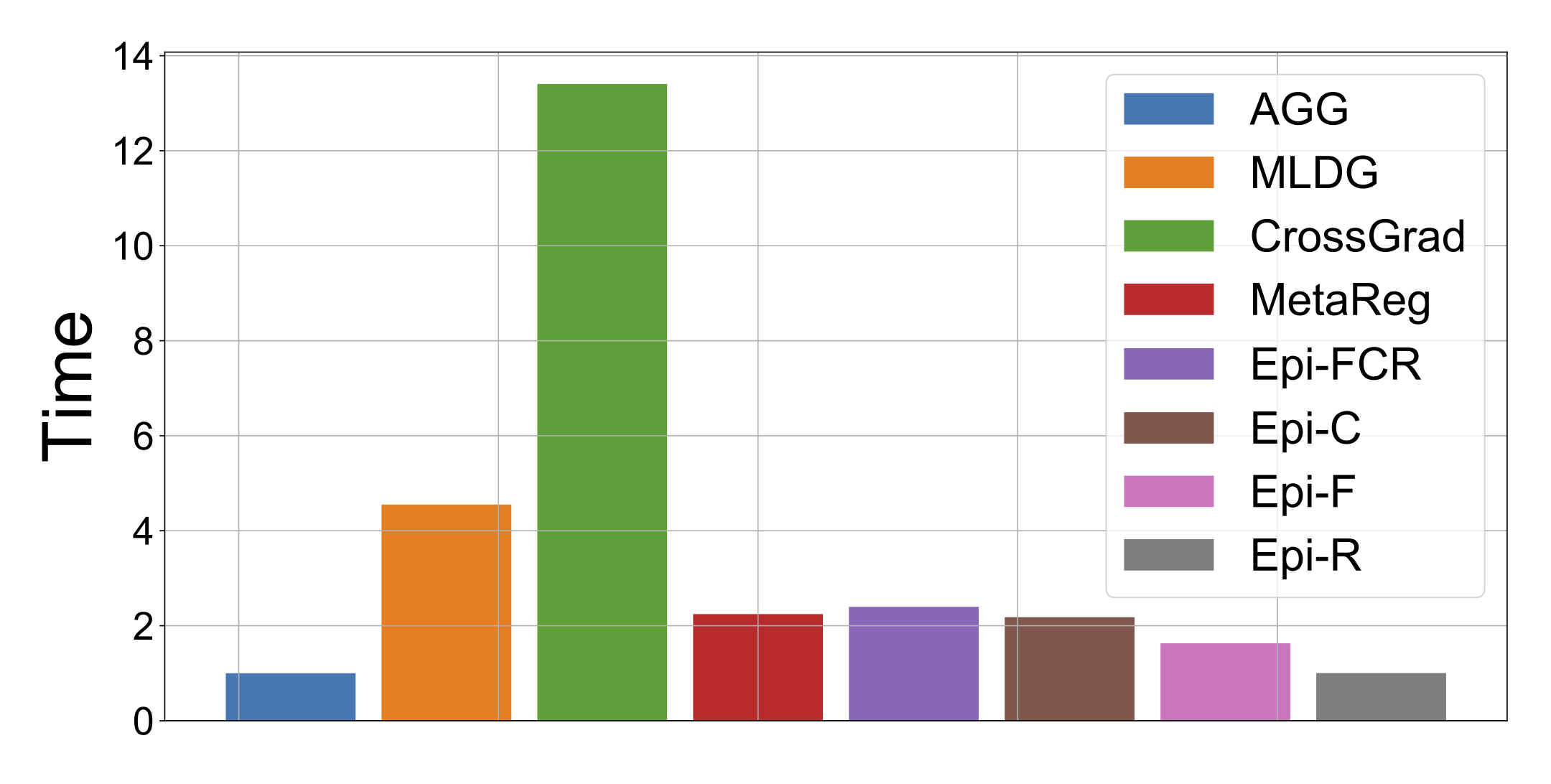}
    \vspace{-0.63cm}
    \caption{}
    \label{fig:comp-cost}
    \end{subfigure}
    \vspace{-0.7cm}
    \caption{\small (a) Ablation study on PACS ($\uparrow$). (b) Computational cost comparison on PACS ($\downarrow$).}
    \vspace{-0.3cm}
\end{figure}

\keypoint{Ablation Study} To understand the contribution of each component of our model, we perform an ablation study using PACS-AlexNet shown in Fig.~\ref{fig:ab-pacs}. Episodic training for the feature extractor, gives a 1.6\% boost over the vanilla AGG. Including episodic training of the classifier, further improves 0.5\%. Finally, combine all the episodic training components, provides 3.3\% improvement over vanilla AGG. This confirms that each component of our model contributes to  final performance. 

\keypoint{Cross-Domain Testing Analysis} To understand how our Epi-FCR method obtains its improved robustness to domain shift, we study its impact on cross-domain testing. Recall that when we activate the episodic training of the agnostic feature extractor and classifier, we benefit from the domain specific branches by routing domain $i$ data across domain $j$ branches. E.g., we feed: $\mathbf{x}_i\mapsto \theta \mapsto \psi_j \mapsto y_i$ to train Eq.~\ref{eq:agg-reg-feat}, and $\mathbf{x}_i\mapsto \theta_j \mapsto \psi \mapsto y_i$ to train Eq.~\ref{eq:agg-reg-clf}. 

Therefore it is natural to evaluate cross-domain testing after training the models. As illustrated in Fig.~\ref{fig:cross-domain-test-acc}\cut{\footnote{To save space we only display the leave-photo-out split. The others are consistent with these observations.}}, we can see that the episodic training strategy indeed improves cross-domain testing performance. For example, when we feed domain $A$ data to domain $C$ classifier $\mathbf{x}_A\mapsto \theta \mapsto \psi_C \mapsto y_A$, the Episodic-trained agnostic extractor $\theta$ improves the performance of the domain-C classifier who has never experienced domain A data (Fig.~\ref{fig:cross-domain-test-acc}, left); and similarly for the Episodic-trained agnostic classifier.  


\keypoint{Analysis of Solution Robustness} In the above experiments we confirmed that our episodic model outperforms the strong AGG baseline in a variety of benchmarks, and that each component of our framework contributes. In terms of analysing the mechanism by which episodic training improves robustness to domain shift, one possible route is through leading the model to find a higher quality minima. Several studies recently have analysed learning algorithm variants in terms of the quality of the minima that they leads a model to \cite{Keskar2017sharpminima,chaudhar2017entropySGD}. 

\begin{figure*}[tb]
    \centering
    \includegraphics[width=0.238\linewidth]{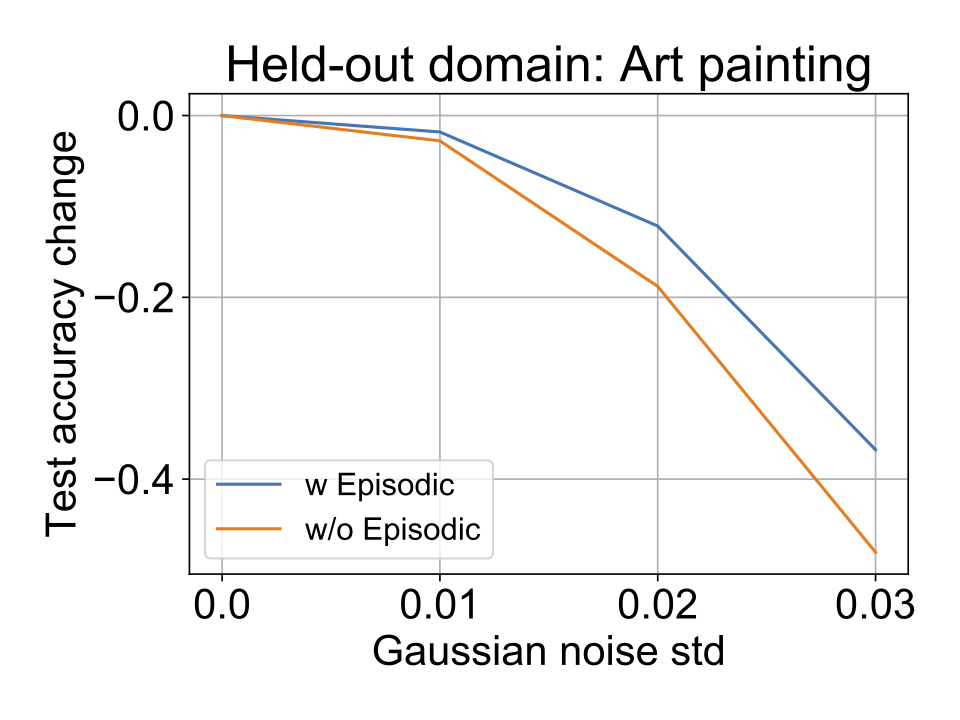}
    \includegraphics[width=0.238\linewidth]{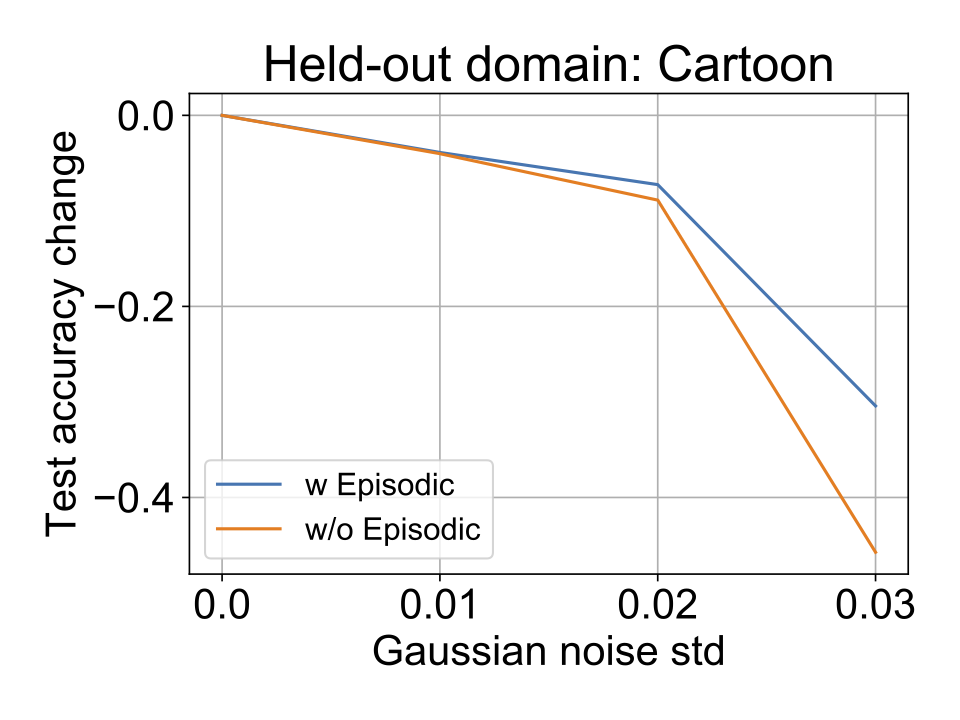}
    \includegraphics[width=0.238\linewidth]{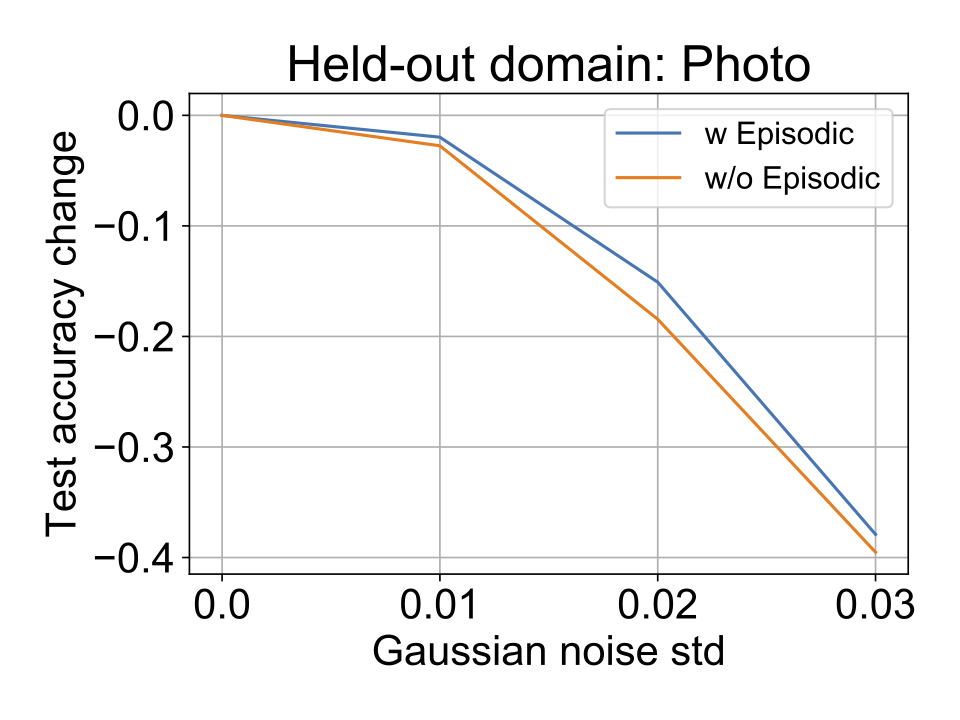}
    \includegraphics[width=0.238\linewidth]{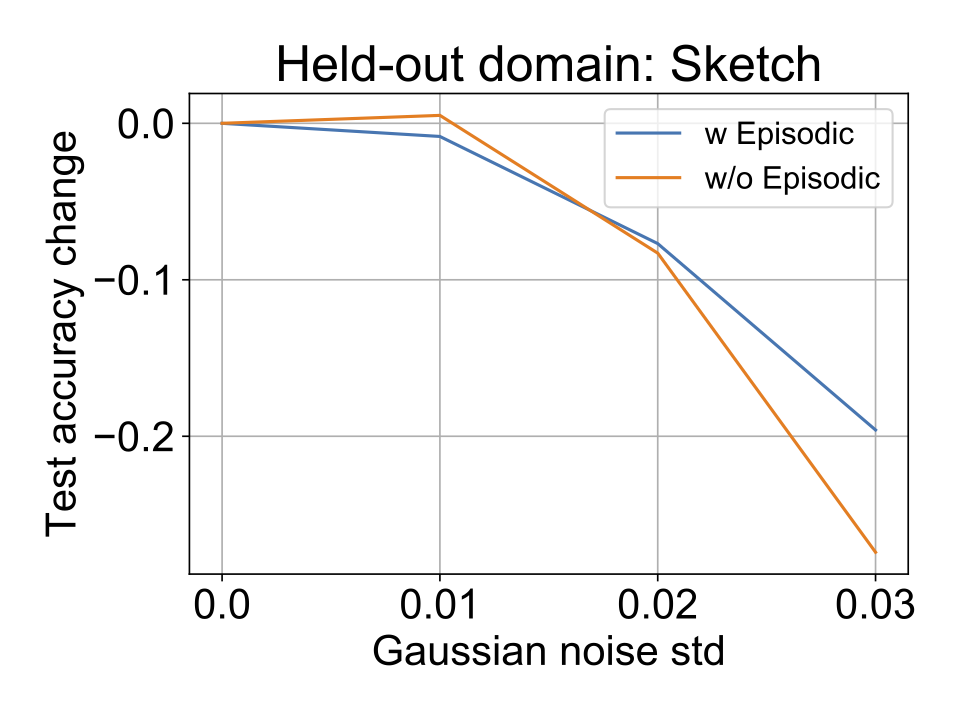}
    \vspace{-0.3cm}
    \caption{\small Minima quality analysis: Episodic training (Epi-FCR) vs baseline (AGG). 
    }
    \label{fig:acc-vs-noise}
\end{figure*}

One intuition is that converging to a `wide' rather than `sharp' minima provides a more robust solution, because perturbations (such as domain shift, in our case) are less likely to cause a big hit to accuracy if the model's performance is not dependent on a very precisely calibrated solution. Following \cite{Keskar2017sharpminima,Zhang2018dml}, we therefore compare the solutions found by AGG and our Epi-FCR by adding noise to the weights of the converged model, and observing how quickly the testing accuracy decreases with the magnitude of the noise. From  Fig.~\ref{fig:acc-vs-noise} we can see that both models' performance drops as weights are perturbed, but our Epi-FCR model is more robust to weight perturbations. This suggests that the minima found by Epi-FCR is a more robust one than that found by AGG, which may explain the improved cross domain robustness of Epi-FCR compared to AGG.

\keypoint{Computational Cost} Our Episodic model is comparable in cost overall to many contemporaries. Our Epi-C variant does require training multiple feature extractors for the source domains (as do \cite{Khosla12undobias,Xu2014lre,Li2017dg,Massimiliano2018ICIP}). However, users are more practically interested in testing performance, where our model is as small, fast and simple as AGG (unlike, e.g., \cite{Xu2014lre,Massimiliano2018ICIP}). In terms of training requirements, we note that only the Epi-C variant requires multiple feature extractor training, so Epi-FR can still safely be used if this is an issue. Furthermore if a large number of source domains are present, we can sample a subset of them at each batch. 

Concretely, we compare the training time of different methods in Fig.~\ref{fig:comp-cost}. All the methods were run on PACS (ResNet-18) for $3k$ iterations with CPU: Intel i7-7820 (@3.60GHz x 16) and GPU: 1080Ti. As expected vanilla AGG is the fastest to train (9.8 mins), so we regard it as the the base unit. The second tier are our Epi-F and Epi-R. As expected without Epi-C, our Epi-F and Epi-R variants run fast. The next tier are MetaReg, Epi-FCR and MLDG. And the most expensive one is CrossGrad. Although the use of `Epi-C' here requires domain-specific feature extractors, our Epi-FCR is still comparably efficient. This is because our episodic training does not generate multi-step graph unrolling or meta-optimization in gradient updates. As a result, our time cost is on par with MetaReg~\cite{NIPS2018_metareg} and faster than MLDG~\cite{Li2018MLDG} and CrossGrad~\cite{shankar2018generalizing}. 



\begin{table*}[t]
    \centering
    \begin{tabular}{cc}
    \scalebox{0.68}{
    \begin{tabular}{c|ccc}\toprule
    \multirow{2}{*}{Setting} & \multicolumn{2}{c}{Updated in Target Domain?} & \multirow{2}{*}{Novel Target Labels?}\\
    & Feature Extractor & Classifier \\
    \midrule
    Homogeneous DG  & N  &  N & N\\
    Heterogeneous DG  & N  & Y & Y\\
    \bottomrule
    \end{tabular}
    }
    &\hspace{1.0cm}
    \scalebox{0.68}{
    \begin{tabular}{c|rcrr}
    \toprule
       Benchmark  & \# of data & \# of Domains & \# of tasks & task space\\ \midrule
       VLCS   & 10,729 & 4 & 5 & Homo.\\
       PACS   & 9,991 & 4 & 7 & Homo. \\
       VD-DG  & 1,521,005  & 10 & 3,128 & Hetero.\\
    \bottomrule
    \end{tabular}
    }
    \end{tabular}
    \vspace{-0.3cm}
    \caption{\small Left: Difference between conventional homogeneous DG setting and new heterogeneous DG setting.  Right: Contrasting the larger scale of our VD-DG (excluding ImageNet) vs previous DG benchmarks.}
    \label{tab:vddg-vs-exdg}
\end{table*}

\begin{table*}[t]
\centering
\scalebox{0.66}{
\begin{tabular}{l|c|ccc|ccc|ccc|ccc|ccc}
\toprule
\multirow{2}{*}{Target} & \multirow{2}{*}{ImageNet PT} & \multicolumn{3}{c|}{MLDG \cite{Li2018MLDG}} & \multicolumn{3}{c|}{CrossGrad~\cite{shankar2018generalizing}} & \multicolumn{3}{c|}{AGG} & \multicolumn{3}{c|}{DANN \cite{ganin2016dann}} & \multicolumn{3}{c}{Epi-R} \\ \cmidrule(){3-17}
 &  & Concat & Mean & Combine & Concat & Mean & Combine & Concat & Mean & Combine & Concat & Mean & Combine & Concat & Mean & Combine \\ \midrule
Aircraft & 12.7 & 17.4 & 14.2 & 15.7& 17.2 & 13.7 &15.9& 17.4 & 14.6 & 15.7 & 17.4 & \textbf{15.0} & \textbf{16.0} & \textbf{17.7} & 13.9 & 15.5 \\
D. Textures & 35.2 & 38.3 & 34.6 &32.5& 34.6 & 31.4 &32.2& 37.7 & 35.1 & 31.5  & 37.9 & 36.6 & 33.0 & \textbf{40.2} & \textbf{37.8} & \textbf{33.9} \\
VGG-Flowers & 48.1 & 54.0 & 53.2 &54.4& 49.2 & 49.3 &54.9& \textbf{56.3} & 52.0 & \textbf{57.0} & 55.5 & 52.2 & 53.7  & 55.4 & \textbf{53.0} & 55.9 \\
UCF101 & 35.0 & 44.4 & 36.7 &34.9& 42.7 & 35.7 &35.2& 43.3 & 35.0 & 36.1 & 44.5 & 36.1 & 33.9 & \textbf{45.7} & \textbf{37.1} & \textbf{37.3} \\ \midrule
Ave. & 32.8 & 38.5 & 34.7 &34.4& 35.9 & 32.5 &34.6& 38.7 & 34.2 & 35.1 & 38.8 & 35.0 & 34.1 & \textbf{39.7} & \textbf{35.5} & \textbf{35.7} \\ \midrule
VD-Score & 185 & 279 & 194 &169& 241 & 169 &169& 265 & 185 & 172 & 277 & 202 & 165 & \textbf{304} & \textbf{217} & \textbf{194} \\ 
\bottomrule
\end{tabular}
}
\vspace{-0.3cm}
\caption{\small Results of top-1 accuracy (\%) and visual decathlon overall scores of different methods on VD-DG. Train on CIFAR-100, Daimler Ped, GTSRB, Omniglot, SVHN, and optionally ImageNet (Combine). Test on Aircraft, D. Textures, VGG-Flowers, UCF101. }
    \label{tab:vd-5s4t-brief}
\end{table*}

\subsection{Evaluation on \textbf{\textit{VD-DG}} dataset}
\keypoint{Heterogeneous Problem Setting}
Visual Decathlon contains ten datasets and was initially proposed as a multi-domain learning benchmark  \cite{Rebuffi17}. We re-purpose Decathlon for a more ambitious challenge of domain generalisation. As explained earlier, our motivation is find out if DG learning can improve the \emph{defacto} standard `ImageNet trained CNN feature extractor' for use as a fixed off-the-shelf representation for new target problems. In this case the feature extractor is trained on the source domain, and used to extract features of the target domain data. Then a target domain-specific classifier (we use SVM) is trained to classify in the target domain. As explained in Table~\ref{tab:vddg-vs-exdg} (left), this is quite different from the standard DG setting in that target domain labels \emph{are} used (for shallow \emph{classifier} training), but the focus here is on the robustness of the  learned feature when generalising to \emph{represent} new domains and tasks without further fine-tuning. If DG training can improve feature generalisation compared to a vanilla ImageNet CNN, this could be of major practical value given the widespread usage of this workflow by vision practitioners.

Besides evaluating a potentially more generally useful problem setting compared to standard homogeneous DG, our VD experiment is also a larger scale evaluation compared to existing DG studies. As shown in Table~\ref{tab:vddg-vs-exdg} (right), VD-DG has twice the domains of VLCS and PACS and is an order of magnitude larger evaluation in terms of data and category numbers. 


\keypoint{Settings}  We consider  five larger datasets in VD (CIFAR-100, Daimler Ped, GTSRB, Omniglot and SVHN\cut{, excluding ImageNet\footnote{We always exploit ImageNet as an initial condition, but do not include it in DG training for computational feasibility}}) as our source domains, and the four smallest datasets (Aircraft, D. Textures, VGG-Flowers and UCF101) as our target domains. The goal is to use DG training among the source datasets to learn a feature which outperforms the off-the-shelf ImageNet-trained CNN that we use as an initial condition. We use ResNet-18 \cite{He2016resnet} as the backbone model, and resize all  images to 64$\times$64 for computational efficiency. To support the VD heterogeneous label space, we assume a shared feature extractor, and a source domain-specific classifier. We perform episodic DG training among the source domains, using our (R)andom classifier model variant, which supports heterogeneous label-spaces. After DG training, the model will then be used as a fixed feature extractor for the held out target domains. \doublecheck{With regards to use of ImageNet during training, we consider two settings: (i) Use ImageNet CNN as initial condition, but exclude ImageNet data from DG training, (ii) Include ImageNet as a sixth source domain for DG training. The former helps to constrain training cost, but loses some performance due to the forgetting effect. Therefore we combine (concatenation and mean-pooling) the original ImageNet pre-trained features with the VD-DG trained features. }
\cut{These are combined by combination (concatenation and mean-pooling) with the original ImageNet pre-trained features\footnote{Since ImageNet is excluded from source domains for computational feasibility, there is loss of performance for all models compared to the original feature due to the forgetting effect.}} In each case the final feature is used to train a linear SVM for the corresponding task, as per common practice. We train the network using the M-SGD optimizer (batch size/per domain=32, lr=1e-3, momentum=0.9, weight decay=1e-4) for 100k iterations where the lr is decayed in 40k, 80k iterations by a factor 10. We set $\lambda_3=\frac{2.5}{t+50}$, $t$ is the iteration num. 

\keypoint{Results} From the results in Table~\ref{tab:vd-5s4t-brief}, we observed that: (i) All methods use the extra data in VD to improve on the initial features (`ImageNet PT'). (ii) In terms of other DG competitors: Only  MLDG, CrossGrad, and DANN were feasible to run on the scale of VD; with others either not supporting heterogeneous label-spaces or scaling to this many domains/examples.  (iii) Our Epi-R improves on the strong AGG baseline and DG competitors in both average accuracy, and also the VD score recommended in preference to accuracy in \cite{Rebuffi17}. This demonstrates the value of our Episodic training in learning a feature that is robust to novel domains. (iv) Our concatenation strategy provided the best overall performance compared to directly including ImageNet as a source domain (`Combine'). This partly due to using a fixed 100k iterations to constrain training time. With enough training, the latter option is likely to be best. Overall this is the first demonstration that any DG method can improve robustness to domain shift in a larger-scale setting, across heterogeneous domains, and make a practical impact in surpassing ImageNet feature performance\footnote{We note one concurrent study of the heterogeneous DG setting \cite{li2019featureCritic} considered the VD-DG benchmark that we propose here. Their results are slightly higher due do use of a larger image size and cross-validation of SVM parameters (we use sklearn defaults).}.

\section{Conclusion} We addressed the domain generalisation problem by proposing a simple episodic training strategy that mimics train-test domain-shift during training, thus improving the trained model's robustness to novel domains. We showed that our method achieves state-of-the-art performance on all the main existing DG benchmarks. We also performed the largest DG evaluation to date, using the Visual Decathlon benchmark. Importantly, we provided the first demonstration of DG's potential value `in the wild' -- by demonstrating our model's potential to improve the performance of the \emph{defacto} standard ImageNet pre-trained CNN as a fixed feature extractor for novel downstream problems. 

\keypoint{Acknowledgements} This work was supported by EPSRC
grant EP/R026173/1.


{\small
\bibliographystyle{ieee_fullname}

}

\clearpage
\newpage
\appendix

\begin{table}[t]
\centering
\scalebox{0.6}{
\begin{tabular}{c|ccccccc}
\toprule
Target  & AGG& Ensemble & Epi-FCR(S2B) & First Blk & Third Blk  & Fixed DSNN & Epi-FCR  \\
\midrule
A. & 77.6 &79.3 & 80.4 & 78.8 & 81.0  & 79.1 & 82.1  \\
C. & 73.9 &75.9 & 75.3 & 75.5 & 75.6  & 76.6 &  77.0 \\
P. & 94.4 &95.4 & 94.4 & 93.7 & 92.2  & 93.5 &  93.9 \\
S. & 70.3 &71.2 & 73.4 & 74.5 & 74.1  & 74.7 &  73.0 \\
\midrule
Ave. & 79.1  &80.4 & 80.9 & 80.6 & 80.7 & 80.9 & 81.5 \\
\bottomrule
\end{tabular}
}
\caption{Further evaluation on PACS using ResNet-18.}
\label{appendix:tab:pacs}
\end{table}

\cut{\begin{table}[t]
\centering
\scalebox{0.7}{
\begin{tabular}{c|cccc}
\toprule
Target &   ImageNetPT & AGG & DANN& Epi-R \\
\midrule
Aircraft  & 12.7 & 15.7 & \textbf{16.0} & 15.5 \\
D. Textures   & \textbf{35.2} & 31.5& 33.0 & 33.9 \\
VGG-Flowers   & 48.1 & \textbf{57.0} &53.7& 55.9 \\
UCF101   & 35.0 & 36.1 &33.9& \textbf{37.3} \\
\midrule
Ave.  & 32.8 & 35.1 & 34.1 &\textbf{35.7} \\
VD-Score  &  185 & 172 & 165& \textbf{194} \\
\bottomrule
\end{tabular}
}
\caption{Results on VD-DG with ImageNet in source domains. Best in bold.}
\label{appendix:tab:vd}
\end{table}}

\section{Additional analysis}
We conduct some analysis to better understand our episodic training method, and its contributions.

\keypoint{Comparison model ensemble} In our current implementation of episodic training, besides the AGG model we regularize, $n$ domain-specific branches are used for generating DG episodes. In this way, it increases the total parameters during training to $n+1$ times that of AGG, although in the end only a single AGG branch is used for testing. To verify that the benefit is not solely due to additional parameters, we compare our episodic-training method with the ensemble of $n+1$ AGG models on PACS using ResNet-18. The result in Table~\ref{appendix:tab:pacs} shows that our episodic training is more effective than the ensemble model. Crucially this is despite the fact that Epi-FCR is $1/(n+1)$th the size of the full ensemble during testing. 

\keypoint{Flexibility of episodic training} a) \emph{Global sharing parameters:} In our current demonstration, we use $n+1$ branches to conduct the episodic training. To reduce the total trainable parameters, we can also globally share the bottom feature layers and episodic-train the rest feature layers and classifier. For example, globally sharing the first two blocks and episodic-training the remaining parameters still leads to a $1.8\%$ improvement over the baseline (see Table~\ref{appendix:tab:pacs}, S2B). b) \emph{Intermediate feature layers:} Applying episodic training around a typical feature vs classifier module split is intuitive, but other options are possible. For example, we evaluate splitting the modules at intermediate feature layers: first block, and third block of ResNet-18 on PACS. The results, in Table~\ref{appendix:tab:pacs} (First Blk and Third Blk), show that episodic training can also work with intermediate layer splits, but the original design is better. c) \emph{Fixed domain-specific branches:} Currently, we train the domain-specific classification losses and the episodic training losses jointly. We also evaluate our episodic training using the pre-trained domain-specific branches. From the result in Table~\ref{appendix:tab:pacs} (Fixed DSNN), we can see that episodic training using fixed domain-specific branches is still effective with $1.8\%$ performance improvement over AGG.

\end{document}